\documentclass[11pt]{report}
\usepackage{times}
\usepackage{latexsym}
\usepackage{url}
\usepackage{graphicx}
\usepackage{multirow}

\title{Question Answering Against Very-Large Text Collections\\Darwin 2008 final Project Report}

\author{
  Leon Derczynski, MComp\\
  Richard Shaw, MCompAI\\
  Ben Solway, MComp\\
  Wang Jun, Msc\\
  {\tt \{aca00lad, aca04rcs, aca04bs, acp07jw\}@sheffield.ac.uk}}

\date{13th May 2008}

\begin{document}
\maketitle
\tableofcontents

\chapter{Introduction}
\section{Question Answering}
Question answering involves developing methods to extract useful information from large collections of documents. This is done with specialised search engines such as Answer Finder. The aim of Answer Finder is to  provide an answer to a question rather than a page listing related documents that may contain the correct answer. So, a question such as ``How tall is the Eiffel Tower" would simply return ``325m" or ``1,063ft".

Our task was to build on the current version of Answer Finder by improving information retrieval, and also improving the pre-processing involved in question series analysis

\section{Overview of aims in Research Proposal}
In our research proposal we aimed to:
\begin{enumerate}
\item {\bf Produce a set of Gold Standard reformulated question series}: 
The gold standard would be created by hand, outlining the best possible question created from a QSA format. Alternative acceptable questions were rated on a sliding scale (from 1 to 10) relating to how useful the reformulation was, with 10 denoting the most useful reformulation. The aim of a scoring system is to reflect an improvement in QSA. It was envisioned that the failure analysis tool could be used to evaluate the scoring system by showing the improvements to the system when entering the different reformulations. The points could then be weighted according to the improvements shown in coverage and redundancy. It was thought a sample of the question series would have to be used as it would be time consuming to build a Gold Standard (GS) for all questions. In creating this GS we would create a number of guidelines which could be followed to produce the best possible reformulation of a question. These methods could then be used to produce gold standard questions from other question series.

\item {\bf Find failures in existing IR systems}:
It was our intention to explore expanding the FMA tool to allow the adjustment of an IR engine's parameters. The system could test multiple candidate configurations of an engine in a single run, thereby allowing the comparison and analysis of individual parameters. This should help give an in-depth look into the impact that any adjustments have.

\item {\bf Explore the possible replacement of Lucene as our IR component}:
Our current IR engine, Lucene, is aging and so we were keen to explore the possibility of replacing it. We had found two possible replacements at the time of writing our research proposal, namely Indri, based on the Lemur toolkit, and Terrier, developed at the University of Glasgow. Our aim was to integrate these with the FMA tool, allowing us to assess them thoroughly and score any improvement over the current Lucene performance. 

\item {\bf Augment our search with complex queries}:
We also planned to explore improving IR engine performance by leveraging the power of the query languages available in some alternative search engines.

\end{enumerate}
\section{Overview of objective changes}
During the resulting research, we made some slight alterations with regard to how we implemented the above proposals.
\begin{enumerate}
\item {\bf Creating the Gold Standard}: We proposed that we would have different levels of reformulation and that we would manually create and assign scores to these levels. To be specific, we needed to measure the similarity of reformulations. This resulted in a gold standard that only listed the very best possible reformulations. This would then act as a benchmark against which other reformulations 
could be assessed.
\item {\bf Information Retrieval}: Increasing the complexity of IR engine queries proved to be unreasonable, as different engines had widely varying strengths of query language. Further, no direction immediately presented itself to guide us in altering queries. Thus, we constructed a data driven approach to query expansion for question answering, and followed this through, producing concrete deliverables. The failure analysis based comparisons didn't change; neither did our selection of platform or engines. 
\end{enumerate} 
\section{Summary of following sections}
Chapter Two will review our progress with regards to developing a Gold Standard and the
development of the metrics required to evaluate reformulations. Chapter Three will address our progress regarding Information Retrieval, including evaluating IR systems for QA and finding difficult questions. We will then present our overall conclusions and identify opportunities for future work in Chapter Four.

\chapter{Improving Question Series Analysis}

\section{Introduction} 

In the research proposal we talked about creating a Gold Standard for question reformulation. We said that we would have different levels of reformulations and that we would manually create and assign scores to these values. We decided that this was too vague and that we needed to create a metric to measure the similarity of the metrics. 

\section{Creating a Gold Standard}

We created a large set of GS questions, but the method by which we created them changed slightly from the original proposal as we decided that our proposed method was not specific enough. We decided that we needed to create a Gold Standard consisting of the best reformulations, and then derive guidelines for creating effective gold standards for other question series. We would measure how effective our reformulations were against un-reformulated question series and those questions reformulated so that the target of the question series were appended to questions.

Instead of taking each question and making several reformulations and rating each one on a sliding scale, we took a subset consisting of 10 of the 65 question series to be reformulated and two of us produced individual variations, creating one or more possible reformulations, before reconciling them.  During this stage we looked at our reformulations and eliminated those which were unsatisfactory and were unlikely to be created automatically. From the remainder we derived a set of guidelines by which we could reformulate the rest of the question series and which would be useful for anyone else wanting to create Gold Standard reformulations. Reformulating another set of 10 question series separately and then reconciling them further refined these guidelines. In total we created a Gold Standard consisting of 448 reformulations, from 406 questions contained within 65 question series. This and other reformulation methods were then evaluated using a suitable metric, as described below in ``Creating A Metric".

\subsection{Gold Standard Guidelines}

Analysis of the Gold Standard reformulations allowed us to create a set of guidelines for creating future Gold Standards, which are summarised below and discussed further in~\cite{shaw:2008:earqqs}. These guidelines should allow anyone to create their own set of Gold Standard reformulations. The emphasis here is on producing guidelines, and not rigid rules as such is the complexity of English that it is not feasible to cover every possible sentence.

\begin{enumerate} 

\item {\bf Context independence and readability}: The reformulation of
questions should be understandable outside of the question series
context. The reformulation should be written as a native speaker would
naturally express it; this means, e.g., that stop words are included.

\item {\bf Reformulate questions so as to maximise search results}:
Example: {\it ``Who was William Shakespeare?"} vs {\it ``Who was
Shakespeare?"}. William should be added to the phrase as it adds
extra information which could allow more results to be found.

\item {\bf Target matches a sub-string of the question}: If the target
string matches a sub-string of the question the target string should
substitute the entirety of the substring. Stop-words should not be
used when determining if strings and target match but should usually
be substituted along with the rest of the target.

\item {\bf Rephrasing}: A Question should not be unnecessarily rephrased.

\item {\bf Previous Questions and Answers}: Questions which include a
reference to a previous question should be reformulated to include a
PREVIOUS\_ANSWER variable. Another reformulation should also be
provided should a system know it needs the answer to the previous question 
but has not found one. This should be a reformulation of the previous question within the current question.

\item {\bf Targets that contain brackets}: Brackets in target should be dealt 
with in the following way. The full target should be substituted into the question in the correct place as one of the Gold Standards. The target without the bracketed word and with it should also be included in the Gold Standard.

\item {\bf Stemming and Synonyms}: Words should not be stemmed and synonyms should not be used unless they are found in the target or the current question series. If they are found then both should be used in the Gold Standard.

\item {\bf It}: The word {\it it} should be interpreted as referring to either the 
answer of the previous question of that set or if no answer available to the target itself.

\item {\bf Pronouns (1)}: If the pronouns {\it he} or {\it she} are used
within a question and the TARGET is of type `Person' then substitute
the TARGET string for the pronoun. If however the PREVIOUS\_ANSWER is
of type `Person' then it should be substituted instead as in this case
the natural interpretation of the pronoun is to the answer of the
previous question.

\item {\bf Pronouns (2)}: If the pronouns {\it his}/{\it hers}/{\it their}
are used within a question and the TARGET is of type `Person' then
substitute the TARGET string for the pronoun appending the string
``s" to the end of the substitution. If however the PREVIOUS\_ANSWER
is of type `Person' then it should be substituted as the natural
interpretation of the pronoun is to the answer of the previous
question.
\end{enumerate}

\subsection{Conclusion}
The aim of creating a set of Gold Standards was to provide a benchmark against which other question reformulation methods could be compared. Though the Gold Standard was sufficiently large to create both a series of helpful guidelines and to achieve the proprosed goals, we feel that expanding the Gold Standard to at least 1000 reformulations will provide a more reliable set of guidelines and be more useful for evaluating reformulation methods.

\section{Creating a Metric}

As described above we decided that we would have only the best reformulation(s) in the Gold Standard instead of the full available gamut as originally planned (including the lowest scoring ones). We also decided that the system should evaluate the closeness of the QA system reformulations on a scale of 0 to 1 (1 being the highest) with 1 indicating an identical match. To compare the Gold standard against the existing QA system reformulations we needed to find a metric that gave good scores to reformulations which seem intuitively closer to the gold standard. 

There are many different systems which attempt to measure string similarity. We experimented with tools like METEOR~\cite{lavie-agarwal:2007:WMT} and ROUGE~\cite{lin:2004:ACLsummarization} but found them unsuitable for this task. The scores returned concentrate too much on the ordering of words. Furthermore both ROUGE and METEOR were developed to compare larger stretches of text -- they are usually used to compare paragraphs rather than sentences. This could be one reason why ordering is valued more highly in this kind of tool. It was also difficult to adjust these tools to give more sensible scores for our reformulations. We decided developing our own metric would be simpler than trying to adapt one of these existing tools.

To explore candidate similarity measures we created a program which would take as input a list of reformulations to be assessed and a list of gold standard reformulations, and compare them using a selection of different string comparison metrics.

To find out which of these metrics best scored reformulations in the way which we desired, we created a set of test reformulations to compare against the gold standard reformulations.

Three test datasets were created: one where the reformulation was simply the original question, one where the reformulation included the target appended to the end, and one where the reformulation was identical to the gold standard. The idea here was that the without target question set should score less than the with target question set and
the identical target question set should have a score of 1 (the highest possible score).  

We then had to choose a set of metrics to test and chose to use
metrics from the SimMetrics library as it is an open source extensible
library of Similarity or Distance Metrics\
\footnote{http://www.dcs.shef.ac.uk/\~sam/simmetrics.html.}.

The next task was to run the metrics provided by SimMetrics over the three datasets and look to see which of the metrics score as we would expect over these datasets. From these results we found that certain metrics were not appropriate. Some of them did the opposite to what we required them to do, in that they scored a reformulation without the target higher than one with the target. This could be due to over-emphasis on word ordering. These metrics were discounted at this stage. Other metrics were also discounted as the difference between ``With Target" and ``Without Target" was not large enough; it would have been difficult to measure improvements in the system with a small difference.

The four best performing metrics were: DiceSimilarity, JaccardSimilarity, BlockDistance and CosineSimilarity. The next task was to refine these metrics so that the scores returned were more useful for the system i.e. the gaps between ``with target", ``without target" and ``identical" are as big as possible. To do this we decided to try and increase the importance of ordering by also taking into account shared bigrams and trigrams. As we do not want ordering to be too important in our metric we introduced a weighting mechanism into the program to allow us to use a weighted combination of shared unigrams, bigrams and trigrams.

We then ran the four best performing metrics over the three datasets with different weightings on bigrams and trigrams. From the results it was clear the both bigrams and trigram reduced the gap between ``with target" and ``without target". Trigrams reduced this gap too much and we decided to not use them in further weightings. Although bigram analysis reduced the ``with target" and ``without target" gap they also extended the gap between ``with target" and ``identical" datasets which is useful. The objective now was to find the best balance of the gaps. We found that a weighting of unigrams 2 and bigrams 1 gave the best results with the metric providing the best score being JaccardSimilarity. 

Using the metrics with our QA system we found that it scored it sensibly. It gave it a score in between the ``with target" and ``identical" datasets which is correct, as our current system did attempt to order the words correctly but sometimes failed and fell back to adding the target on the end of the reformulation. 

\subsection{Conclusion}

The aim of creating this metric was to quickly evaluate the pre-processing code in the QA system against a gold standard. We were successful in finding a useful metric and bigram weighting for this metric. Further work could be to refine the weighting used further using machine learning techniques. We could also automate the process of checking the changes made in the pre-processing code more so that it can be evaluated more easily. This could include highlighting reformulations which improved the most and decreased the most from the changes made in the code.

\chapter{Improving IR}

\section{Introduction}

The IR component of the original Sheffield QA system effectively capped the proportion of corectly answerable questions at around 60{\%}. In our research proposal, we suggested improving the performance of the IR component so that the overall accuracy of the question answering system might exceed this limit.

\section{Evaluating IR Systems for QA}

One of our suggestions was to replace the IR component. The prior Sheffield QA system only used Lucene for IR in the TREC test. The Failure Analysis Tool was previously used to evaluate the performance of the Lucene engine, so we changed this tool (through adding database parameters) to make it be able to support multiple different IR engine configurations. Finally, the Failure Analysis Tool supported all three IR engines, Lucene, Indri and Terrier, which were used in our experiment. It reported on all engines at once and provided comparisons between them.

Two powerful IR engines, Indri and Terrier, had been recommended. We integrated these two IR engines into the Sheffield QA system using its plug-in interface. Another part of this QA system (GetRelevantDocuments) was used to test engines with TREC question sets.

In our experiment, in order to evaluate the performance of three IR engines, we looked at both passage- and document-level retrieval. The original QA system answered questions using Lucene based on passage-level indexes; we then changed programs to add document-level indexes for Lucene based on the AQUAINT corpus. Indri natively supports document-level indexing of TREC format corpora. Passage-level retrieval was done by defining delimiters, which were assigned to the paragraph tags in the corpus; this allows both passage- and document-level retrieval from the same index, set in the query. Terrier doesn�t natively support passage-level indexing or retrieval, so we built a program to chunk original documents in the AQUAINT corpus to passage-level documents then built indexes based on the chunked documents to get passage-level indexes for Terrier. In addition, we directly built document-level indexes for Terrier by using its TREC Terrier application.

The top 20 documents for each question in the TREC2004, TREC2005 and TREC2006 were retrieved using three IR engines at passage level. In addition, the top 20 documents for each question in each TREC year were also retrieved using Indri and Terrier at document level, and we have presented all results in our journal paper~\cite{wang:2008:ddaqeqa}. We didn�t retrieve documents using Lucene at document level over TREC question sets from each year because it was a very time consuming process. For each question set, it usually took 7-10 hours to finish it. Another important reason why we didn�t do it is that passage-level retrieval performs better than document-level retrieval for answer extraction because the amount of noise is somewhat reduced when using passage-level retrieval~\cite{Roberts:Passage}. From the results of our experiment, we can see that the performance of three IR engines doesn�t highlight a strong difference. Indri performs slightly better than other two engines with lenient matching. Moreover, the coverage of all three engines didn�t exceed 64{\%} with strict matching, so simply switching to an alternative IR component won't fix the problem.

Another suggestion is to find failures in existing systems. To do failure analysis, we suggested grouping tested questions by type or their expected answer (EAT). We didn�t do this because we realized that the coverage probably won�t go much over 60{\%} with any engine without making changes to the query strings. As a result, we began to focus on identifying problem questions and attempted to expand them with words extracted from answer passages. We then re-tested expanded questions to see if the overall question answering accuracy improved.

Finally, we suggested augmenting search with complex queries. We didn�t do it because although Indri has a very rich query language, query languages in Terrier and Lucene are not strong, so that it is difficult to do a comparison among three IR engines and we didn�t want to be bound to a single IR engine. Our aim is to find a common approach to improve the performance of answering questions for all possible IR engines used in our system. So we began to focus on finding a data driven approach to query extension.

\section{Finding Difficult Questions}

Given that coverage is so low, there must be a significant number of questions that are difficult. We planned to identify these, hoping further examination could shed light on the problems. Using the failure analysis database, we automatically selected difficult questions and reported lists of these for each IR engine. This provided a sample of data describing situations in which the IR component behaves badly.

To capialise on this data and attempt to improve IR performance, we attempted to identify words that could be used as query extensions. Our hypothesis was that words found in answer-bearing paragraphs would boost performance when used as query extensions. To this end, for each question we found answer-bearing paragraphs based on TREC answer specifications, extracted potentially useful words from them, and built a list of candidate extensions.

Not all documents containing answers are noted, only those checked by the NIST judges ~\cite{bilottiadvice}. Match judgements are incomplete, leading to the potential generation of false negatives, where a correct answer is found with complete supporting information, but as the information has not been manually flagged, the system will mark this as a failure. Assessment methods are fully detailed in Dang et al.~\cite{dang2006otq}. 

A mean redundancy is also calculated over a selection of redundancy measures. Questions with a low mean redundancy are found to be \emph{difficult}, and subjected to further analysis. Filtering the question set by these difficult questions produces a TREC-format question set for re-testing the IR component.

We found that difficult questions did vary between engines and configurations. The size of the difficult question set was not always enough to provide useful data, and so we attempted to increase it, by restricing the way in which difficult questions were determined; sometimes $n$ would be reduced, or only strict measures used, or the number of engines restricted. This successfully increased the answer set. Our full methodoly and results are described in by Derczynski et al.~\cite{wang:2008:ddaqeqa}.

These extended question were then tested, looking for which extension keywords provided the most assistance.

\section{Relevance Feedback}
Some data that can be obtained without prior knowledge of the answers ought to be tested. Knowing which words assist, given hindsight, is of no use unless we can find a technique for query expansion that selects some of these useful words for unseen questions. To test a simple query expansion method, we picked blind relevance feedback (assisted relevance feedback would be unsuitable given the nature of the TREC QA tasks). This operated by attempting an initial retrieval using the base question, and harvesting keywords from the top $r$ documents. Words were ranked according to their term frequency (TF) and then appended to questions to form new query strings.

\subsection{Measuring extension approaches}
\begin{table}
\begin{center}
\begin{tabular}{ | l | r | r | r |}
\hline
 & 2004 & 2005 & 2006 \\
\hline
HEW found in IRT & 4.17{\%} & 18.58{\%} & 8.94{\%} \\
IRT containing HEW & 10.00{\%} & 33.33{\%} & 34.29{\%} \\
RF words in HEW & 1.25{\%} & 1.67{\%} & 5.71{\%} \\
\hline
\end{tabular}
\caption{HEW = ``helpful extension words", the set of extensions that, when added to the query, move redundancy above zero. $r = 5$, $n = 20$, using Indri at passage level. (IRT = ``initially retrieved texs")}
\end{center}
\label{hewrfintersection}
\end{table}

\begin{table}
\begin{center}
\begin{tabular}{ | c || l | l | l | l || l |}
\hline
 & \multicolumn{4}{ | c | }{$r$} & \\
\cline{2-5}
 & \multicolumn{2}{ | c | }{5} & \multicolumn{2}{ | c | }{50} & Baseline \\
\cline{1-5}
Rank & Doc & Para & Doc & Para & \\
\hline
5 & 0.253 & 0.251 & 0.240 & 0.179 & 0.312 \\
10 & 0.331 & 0.347 & 0.331 & 0.284 & 0.434 \\
20 & 0.438 & 0.444 & 0.438 & 0.398 & 0.553 \\
50 & 0.583 & 0.577 & 0.577 & 0.552 & 0.634 \\
\hline
\end{tabular}
\caption{Coverage (strict) using blind RF. Both document- and paragraph-level retrieval used to determine RF terms.\label{brf2k6}}
\end{center}
\end{table}

Using five words for relevance feedback, we measured the intersection between helpful extension words and those chosen for relevance feedback. Our findings (see Table~\ref{hewrfintersection}) suggested that while a noticeable proportion of texts contained potentially helpful words, these words were not often selected by the TF ranking mechanism, which managed to actually find helpful words less than 10{\%} of the time (the other 90{\%} of the time, the added words were simply noise).

This finding was reinforced by the actual performance of TF-based RF, shown in Table~\ref{brf2k6}, where a consistent drop in coverage was found.

\section{Conclusion}

In conclusion we have added Indri and Terrier to the Sheffield QA system and did a comparison for answering TREC question sets of three years among three IR engines, Lucene, Indri and Terrier, based on both passage-level and document-level retrievals. We didn�t find a strong difference at the performance of answering questions among three IR engines and all of them didn�t exceed 60{\%} accuracy.

Further, we decided to focus on expanding query through using a data driven approach to attempt to improve the performance of IR. We successfully identified difficult questions, and then developed a means of extracting extension words, as planned. Our hypothesis that these extension words would help provide answers for difficult questions was then sustained.

We further managed to develop a demonstrably accurate tool for assessing query expansion approaches. This was used to test blind relevance feedback based on term frequency, which it predicted would perform badly. In real tests, blind relevance feedback provided an overall negative effect on feedback.

\chapter{Evaluation}
\section{Conclusion}

The project seems to have been an overall success. Solid work and progress was made in most areas that we planned to tackle, leading to two conference paper submissions.

\subsection{Outcome with respect to proposed work}
With regard to the IR track, we analysed failures in the existing system, though did not attempt to perform any kind of EAT-oriented analysis. Instead, we built a reference list of difficult questions and a body of helpful words, and then made these difficult questions somewhat approachable. We did manage to construct a set of easily interchangeable IR component configurations, measure performance with them, and used all of these plugins in investigating difficult questions. Augmentation of queries didn't happen in the way we imagined - instead of using the native query language of each engine, which would have bound us strongly to specific software, we explored query expansion based on existing data.

Our aim with the QSA track was to create a Gold Standard by hand and a metric for evaluating reformulations. Our original aims were modified when we realised our proposed method of creating Gold Standards was too vague. We instead produced a Gold Standard of only the best possible reformulations and from these derived a set of guidelines that anyone will be able to follow in order to create their own Gold Standard. We were successful in evaluating several metrics and creating a metric to evaluate the pre-processing code in the QA system against the Gold Standard together with appropriate metric and bigram weighting.

\pagebreak
\subsection{Deliverables}

The QSA track produced:

\renewcommand{\labelitemi}{$\star$}
\begin{itemize}
\item A set of 65 Gold Standard reformulated question series.
\item Guidelines for creating Gold Standard reformulations from other question series.
\item Creating a metric to quickly evaluate the pre-processing code in the QA system against a Gold Standard and a useful metric and bigram weighting for this metric.
\item Paper submission to IR4QA 2008~\cite{shaw:2008:earqqs}. 
\end{itemize}

The IR track produced:

\begin{itemize}
\item A tool for predicting the performance of query expansion techniques.
\item Methods for identifying difficult questions and candidate extension terms.
\item A review of IR components for QA.
\item Paper submission to IR4QA 2008~\cite{wang:2008:ddaqeqa}.
\end{itemize}

\section{Future Work}

Further work with regard to QSA will aim to expand the Gold Standard to at least 1000 questions, refining the guidelines as required. The eventual goal is to incorporate the approach into an evaluation tool such that a developer would have a convenient way of evaluating any question reformulation strategy against a large gold standard. One also needs to develop methods for observing and measuring the effect of question reformulation within question preprocessing upon the performance of downstream components in the QA system, such as document retrieval.

An alternative data driven approach to improving IR for QA would be to build associations between recurrently useful terms given question content. Question texts could be stripped of stopwords and proper nouns, and a list of HEWs associated with each remaining term. To reduce noise, the number of times a particular extension has helped a word would be counted. Given sufficient sample data, this would provide a reference body of HEWs to be used as an aid to query expansion. Our query expansion assessment tool needs a more thorough roadtest, and could be better validated. Also, it is possible to build models of the types of words (using e.g. part of speech) useful to each type of question (with e.g. EAT classification), and adjust the words used for query expansion accordingly.

\section*{Acknowledgements}

Special thanks to Mark A. Greenwood and Rob Gaizauskas for their relentless support of our work.

\bibliographystyle{alpha}
\bibliography{qaavltc}

\end{document}